\documentclass[final,twocolumn,a4paper]{article}

\usepackage{graphicx}
\usepackage{amssymb}
\usepackage{times}

\usepackage{natbib}
\usepackage{sectsty}
\sectionfont{\centering\large}

\usepackage{lineno}
\usepackage{multirow}
\usepackage{here}

\title{ Road Damage Detection Using Deep Neural Networks with Images Captured Through a Smartphone}

\author{Hiroya Maeda\footnote{maedahi@iis.u-tokyo.ac.jp}, Yoshihide Sekimoto, Toshikazu Seto, Takehiro Kashiyama, Hiroshi Omata\\
University of Tokyo, 4-6-1 Komaba, Tokyo, Japan}

\date{} %remove date

\begin{document}
\maketitle

\textbf{Abstract:} \textit{Research on damage detection of road surfaces using image processing techniques has been actively conducted, achieving considerably high detection accuracies.
Many studies only focus on the detection of the presence or absence of damage. However, in a real-world scenario, when the road managers from a governing body need to repair such damage, they need to clearly understand the type of damage in order to take effective action. In addition, in many of these previous studies, the researchers acquire their own data using different methods. Hence, there is no uniform road damage dataset available openly, leading to the absence of a benchmark for road damage detection.
This study makes three contributions to address these issues.
First, to the best of our knowledge, for the first time, a large-scale road damage dataset is prepared. This dataset is composed of 9,053 road damage images captured with a smartphone installed on a car, with 15,435 instances of road surface damage included in these road images. In order to generate this dataset, we cooperated with 7 municipalities in Japan and acquired road images for more than 40 hours. These images were captured in a wide variety of weather and illuminance conditions. In each image, we annotated the bounding box representing the location and type of damage.
Next, we used a state-of-the-art object detection method using convolutional neural networks to train the damage detection model with our dataset, and compared the accuracy and runtime speed on both, using a GPU server and a smartphone. Finally, we demonstrate that the type of damage can be classified into eight types with high accuracy by applying the proposed object detection method.
The road damage dataset, our experimental results, and the developed smartphone application used in this study are publicly available
(https://github.com/sekilab/RoadDamageDetector/).}

%\begin{keyword}
%Infrastructure Maintenance \sep Road Damage Detection \sep Object Detection \sep Deep Learning \sep Dataset


%\end{keyword}

%\end{frontmatter}

\section{Introduction}
\label{S:1}
During the period of high economic growth in Japan from 1954 to 1973, infrastructure such as roads, bridges, and tunnels were constructed extensively; however, because many of these were constructed more than 50 years ago~\citep*{MLIT2016}, they are now aged, and the number of structures that are to be inspected is expected to increase rapidly in the next few decades.
In addition, the discovery of the aged and affected parts of infrastructure has thus far depended solely on the expertise of veteran field engineers.
However, owing to the increasing demand for inspections, a shortage of field technicians (experts) and financial resources has resulted in many areas.
In particular, the number of municipalities that have neglected conducting appropriate inspections owing to the lack of resources or experts has been increasing~\citep*{kazuya2013effective}. The United States also has similar infrastructure aging problems~\citep*{aaoshat2008bridging}. The prevailing problems in infrastructure maintenance and management are likely to be experienced by countries all over the world. Considering this negative trend in infrastructure maintenance and management, it is evident that efficient and sophisticated infrastructure maintenance methods are urgently required.

In response to abovementioned problem, many methods to efficiently inspect infrastructure, especially road conditions, have been studied, such as methods using laser technology or image processing. Moreover, there are quite a few studies using neural networks for civil engineering problems in the 11 years from 1989 to 2000~\citep*{adeli2001neural}.
Furthermore, recently, computer vision and machine learning techniques have been successfully applied to automate road surface inspection~\citep*{chun2015asphalt,zalama2014road,jo2015pothole}. 
% add more papers!!

However, thus far, with respect to methods of inspections using image processing, we believe these methods suffer from three major disadvantages:

1. There is no common dataset for a comparison of results; in each research, the proposed method is evaluated using its own dataset of road damage images.
Motivated by the field of general object recognition, wherein large common datasets such as ImageNet~\citep*{deng2009imagenet} and PASCAL VOC~\citep*{everingham2010pascal,everingham2015pascal} exist, we believe there is a need for a common dataset on road scratches.

2. Although current state-of-the-art object detection methods use end-to-end deep learning techniques, no such method exists for road damage detection. 

3. Though road surface damage is distinguished into several categories (in Japan, eight categories according to the Road Maintenance and Repair Guidebook 2013~\citep*{pavement01}), 
many studies have been limited to the detection or classification of damage in only the longitudinal and lateral directions~\citep*{chun2015asphalt,zalama2014road,zhang2016road, akarsu2016fast,maeda2016lightweight}.
Therefore, it is difficult for road managers to apply these research results directly in practical scenarios.

Considering the abovementioned disadvantages, in this study, we develop a new, large-scale road damage dataset, and then train and evaluate a damage detection model that is based on the state-of-the-art convolutional neural network (CNN) method.

The contributions of this study are as follows.

1.We created and released 9,053 road damage images containing 15,435 damages.
The dataset contains the bounding box of each class for the eight types of road damage.
Each image is extracted from an image set created by capturing pictures of a large number of roads obtained using a vehicle-mounted smartphone. The 9,053 images of the dataset contain a wide variety of weather and illuminance conditions. In addition, in assessing the type of damage, the expertise of a professional road administrator was employed, rendering the dataset considerably reliable.

2. Using our developed dataset, we have evaluated the state-of-the art object detection method based on deep learning and made benchmark results. All the trained models are also publicly available on our website\footnote{https://github.com/sekilab/RoadDamageDetector/}. 

3. Furthermore, we showed that the type of damage from among the eight types can be identified with high accuracy.

The rest of the paper is organized as follows. In Section~\ref{RelatedWork}, we discuss the related works. Details of our new dataset are presented in Section~\ref{OurDataset}. The experimental settings are explained in Section~\ref{ExperimentalSetup}. Then, the results are provided in Section~\ref{Results}. Finally, Section~\ref{Conclusions} concludes the paper.

\section{Related Works}
\label{RelatedWork}

\subsection{Road Damage Detection}
\label{RoadCrackDetection}
Road surface inspection is primarily based on visual observations by humans and quantitative analysis using expensive machines.

Among these, the visual inspection approach not only requires experienced road managers, but also is time-consuming and expensive. Furthermore, visual inspection tends to be inconsistent and unsustainable, which increases the risk associated with aging road infrastructure. Considering these issues, municipalities lacking the required resources do not conduct infrastructure inspections appropriately and frequently, increasing the risk posed by deteriorating structures.

In contrast, quantitative determination based on large-scale inspection, such as using a mobile measurement system (MMS)~\citep*{MMS} or laser-scanning method~\citep*{yu2011pavement} is also widely conducted. An MMS obtains highly accurate geospatial information using a moving vehicle; this system comprises a global positioning system (GPS) unit, an internal measurement unit, digital measurable images, a digital camera, a laser scanner, and an omnidirectional video recorder. Though quantitative inspection is highly accurate, it is considerably expensive to conduct such comprehensive inspections especially for small municipalities that lack the required financial resources.

Therefore, considering the abovementioned issues, several attempts have been made to develop a method for analyzing road properties by using a combination of recordings by in-vehicle cameras and image processing technology to more efficiently inspect a road surface.
For example, a previous study proposed an automated asphalt pavement crack detection method using image processing techniques and a naive Bayes-based machine-learning approach~\citep*{chun2015asphalt}. In addition, a pothole-detection system using a commercial black-box camera has been previously proposed~\citep*{jo2015pothole}. In recent times, it has become possible to quite accurately analyze the damage to road surfaces using deep neural networks~\citep*{zhang2016road,maeda2016lightweight,zhang2017automated}. For instance, Zhang et al.~\citep*{zhang2017automated} introduced CrackNet, which predicts class scores for all pixels. However, such road damage detection methods focus only on the determination of the existence of damage. Though some studies do classify the damage based on types---for example, Zalama et al.~\citep*{zalama2014road} classified damage types vertically and horizontally, and Akarsu et al.~\citep*{akarsu2016fast} categorized damage into three types, namely, vertical, horizontal, and crocodile---most studies primarily focus on classifying damages between a few types. Therefore, for a practical damage detection model for use by municipalities, it is necessary to clearly distinguish and detect different types of road damage; this is because, depending on the type of damage, the road administrator needs to follow different approaches to rectify the damage.

Furthermore, the application of deep learning for road surface damage identification has been proposed by few studies, for example, studies by Maeda et al.~\citep*{maeda2016lightweight} and Zhang et al.~\citep*{zhang2016road}. However, the method proposed by Maeda et al.~\citep*{maeda2016lightweight}, which uses 256 $\times$ 256 pixel images, identifies the damaged road surfaces, but does not classify them into different types. In addition, the method of Zhang et al.~\citep*{zhang2016road} identifies whether damage occurred exclusively using a 99 $\times$ 99 patch obtained from a 3264 $\times$ 2448 pixel image. Further, a 256 $\times$ 256 pixel damage classifier is applied using a sliding window approach~\citep*{felzenszwalb2010object} for 5,888 $\times$ 3,584 pixel images in order to detect cracks on the concrete surface~\citep*{cha2017deep}. In these studies, classification methods are applied to input images and damage is detected. Recently, it has been reported that object detection using end-to-end deep learning is more accurate and has a faster processing speed than using a combination of classification methods; this will be discussed in detail in~\ref{ObjectDetectionSystem}. As an example of a method using end-to-end deep learning performing better than tradition methods, white line detection based on end-to-end deep learning using OverFeat~\citep*{sermanet2013overfeat} outperformed a previously proposed empirical method~\citep*{huval2015empirical}. However, to the best of our knowledge, no example of the application of end-to-end deep learning method for road damage detection exists. It is important to note that classification refers to labeling an image rather than an object, whereas detection means assigning an image a label and identifying the object’s coordinates as exemplified by the ImageNet competition~\citep*{deng2009imagenet}.

Therefore, considering this, we apply the end-to-end object detection method based on deep learning to the road surface damage detection problem, and verify its detection accuracy and processing speed. In particular, we examine whether we can detect eight classes of road damage by applying state-of-the-art object detection methods (discussed later in~\ref{ObjectDetectionSystem}) with the newly created road damage dataset (explained in Section~\ref{OurDataset}). Although many excellent methods have been proposed, such as segmentation of cracks on the concrete surface~\citep*{o2014regionally,nishikawa2012concrete}, our research uses an object detection method.

\subsection{Image Dataset of Road Surface Damage}
Though an image dataset of the road surface exists, called the kitti dataset~\citep*{geiger2013vision}, it is primarily used for applications related to automatic driving.
However, to the best of our knowledge, no dataset tagged for road damage exists in the field.
In all the studies focusing on road damage detection described in~\ref{RoadCrackDetection}, in each study, the researchers independently propose unique methods using acquired road images. Therefore, a comparison between the methods presented in these studies is difficult.

Furthermore, according to Mohan et al.~\citep*{MOHAN2017}, there are few studies that construct damage detection models using real data, and 20 of these studies use road images taken directly from above the road. In fact, it is difficult to reproduce the road images taken directly from above the roads, because doing so involves installing a camera outside the car body, which, in many countries, is a violation of the law; in addition, it is costly to maintain a dedicated car solely for road images.
Therefore, we have developed a dataset of road damage images using the road images captured using a smartphone on the dashboard of a general passenger car; in addition, we made this dataset publicly available. Moreover, we show that road surface damage can be detected with considerably high accuracy even with images acquired by employing such a simple method.

\subsection{Object Detection System}
\label{ObjectDetectionSystem}
In general, for object detection, methods that apply an image classifier to an object detection task have become mainstream;
these methods entail varying the size and position of the object in the test image, and then using the classifier to identify the object.
The sliding window approach is a well-known example~\citep*{felzenszwalb2010object}.
In the past few years, an approach involving the extraction of multiple candidate regions of objects using region proposals as typified by R-CNN, then making a classification decision with candidate regions using classifiers has also been reported~\citep*{girshick2014rich}. However, the R-CNN approach can be time consuming because it requires more crops, leading to significant duplicate computation from overlapping crops. This calculation redundancy was solved using a Fast R-CNN~\citep*{girshick2015fast}, which inputs the entire image once through a feature extractor so that crops share the computation load of feature extraction. As described above, image processing methods have historically developed at a considerable pace. In our study, we primarily focus on four recent object detection systems: the Faster R-CNN~\citep*{ren2015faster}, the You Look Only Once (YOLO)~\citep*{redmon2016you,redmon2016yolo9000} system, the Region-based Fully Convolutional Networks (R-FCN) system~\citep*{dai2016r}, and the Single Shot Multibox Detector (SSD) system~\citep*{liu2016ssd}.

\subsubsection{Faster R-CNN}
The Faster R-CNN~\citep*{ren2015faster} has two stages for detection. In the first stage, images are processed using a feature extractor (e.g., VGG, MobileNet) called the~\textit{Region Proposal Network (RPN)} and simultaneously, some intermediate level layers (e.g., "conv5") are used to predict class bounding box proposals. 

In the second stage, these box proposals are used to crop features from the same intermediate feature map, which are subsequently input to the remainder of the feature extractor in order to predict a class label and its bounding box refinement for each proposal.
It is important to note that Faster R-CNN does not crop proposals directly from the image and re-runs crops through the feature extractor, which would lead to duplicated computations. 

\subsubsection{YOLO}
YOLO is an object detection framework that can achieve high mean average precision (mAP) and speed~\citep*{redmon2016you, redmon2016yolo9000}.
In addition, YOLO can predict the region and class of objects with a single CNN.
An advantageous feature of YOLO is that its processing speed is considerably fast because it solves the problem as a mere regression, detecting objects by considering background information. The YOLO algorithm outputs the coordinates of the bounding box of the object candidate and the confidence of the inference after receiving an image as input. 

\subsubsection{R-FCN}
R-FCN is another object detection framework, which was proposed by Dai et al.~\citep*{dai2016r}. Its architecture is that of a region-based, fully convolutional network for accurate and efficient object detection. Although Faster R-CNN is several times faster than Fast R-CNN, the region-specific component must be applied several hundred times per image. Instead of cropping features from the same layer where the region proposals are predicted like in the case of the Faster R-CNN method, in the R-FCN method, crops are taken from the last layer of the features prior to prediction. This approach of pushing cropping to the last layer minimizes the amount of per-region computation that must be performed. Dai et al.~\citep*{dai2016r} showed that the R-FCN model (using Resnet 101) could achieve accuracy comparable to Faster R-CNN often at faster running speeds.

\subsubsection{SSD}
SSD~\citep*{liu2016ssd} is an object detection framework that uses a single feed-forward convolutional network to directly predict classes and anchor offsets without requiring a second stage per-proposal classification operation. The key feature of this framework is the use of multi-scale convolutional bounding box outputs attached to multiple feature maps at the top of the network.

\subsection{Base Network}
\label{baseNetwork}
In all these object detection systems, a convolutional feature extractor as a base network is applied to the input image in order to obtain high-level features. The selection of the feature extractor is considerably important because the number of parameters and layers, the type of layers, and other properties directly affect the performance of the detector. We have selected seven representative base networks, which are explained in~\ref{baseNetwork}, and three base networks to evaluate the results in Section~\ref{Results}. The six feature extractors we have selected are widely used in the field of computer vision. 

\subsubsection{darknet-19}
Darknet-19~\citep*{redmon2016yolo9000} is a base model of the YOLO framework. The model has 19 convolutional layers and 5 maxpooling layers.

\subsubsection{VGG-16}
VGG 16~\citep*{simonyan2014very} is a CNN with a total of 16 layers consisting of 13 convolution layers and 3 fully connected layers proposed in the ImageNet Large Scale Visual Recognition Challenge (ILSVRC) in 2014. This model achieved good results in ILSVRC and COCO 2015 (classification, detection, and segmentation) considering the depth of the layers.

\subsubsection{Resnet}
Resnet, which refers to Deep Residual Learning,~\citep*{he2016deep}, is a structure for deep learning, particularly for CNNs, that enables high-precision learning in a very deep network; it was released by Microsoft Research in 2015. Accuracy beyond human ability is obtained by learning images with 154 layers. Resnet achieved an error rate of 3.57\% with the ImageNet test set and won the first place in ILSVRC 2015 classification task.

\subsubsection{Inception V2}
Inception V2~\citep*{ioffe2015batch} and Inception V3~\citep*{szegedy2016rethinking} enable one to increase the depth and breadth of the network without increasing the number of parameters or the computational complexity by introducing so-called inception units.

\subsubsection{Inception Resnet}
Inception Resnet V2~\citep*{szegedy2017inception} improves recognition accuracy by combining both residual connections and Inception units effectively.

\subsubsection{MobileNet}
MobileNet~\citep*{howard2017mobilenets} has been shown to achieve an accuracy comparable to VGG-16 on ImageNet with only 1/30th of the computational cost and model size. MobileNet is designed for efficient inference in various mobile vision applications. Its building blocks are
depthwise separable convolutions that factorize a standard convolution into a depthwise convolution and a 1 $\times$ 1 convolution, effectively reducing both the computational cost and number of parameters.

\section{Proposed Dataset}
\label{OurDataset}

In this section, we describe our proposed new dataset, including how the data was obtained, how it was annotated, its contents, and issues related to privacy.

\subsection{Data Collection}

Thus far, in the study of damage detection on the road surface, images are either captured from above the road surface or using on-board cameras on vehicles.
When models are trained with images captured from above, the situations that can be applied in practice are limited, considering the difficulty of capturing such images.
In contrast, when a model is constructed with images captured from an on-board vehicle camera, it is easy to apply these images to train the model for practical situations. For example, using a readily available camera like on smartphones and general passenger cars, any individual can easily detect road damages by running the model on the smartphone or by transferring the images to an external server and processing it on the server. 

We selected seven local governments in Japan\footnote{Ichihara city, Chiba city, Sumida ward, Nagakute city, Adachi city, Muroran city, and Numazu city.} and cooperated with the road administrators of each local government to collect 163,664 road images\footnote{We traveled through every municipality covering approximately 1,500 km in total}. Seven municipalities have snowy areas and urban areas that are very diverse in terms of regional characteristics such as the weather and fiscal constraints. 

\begin{figure}[h]
\centering\includegraphics[width=0.6\linewidth]{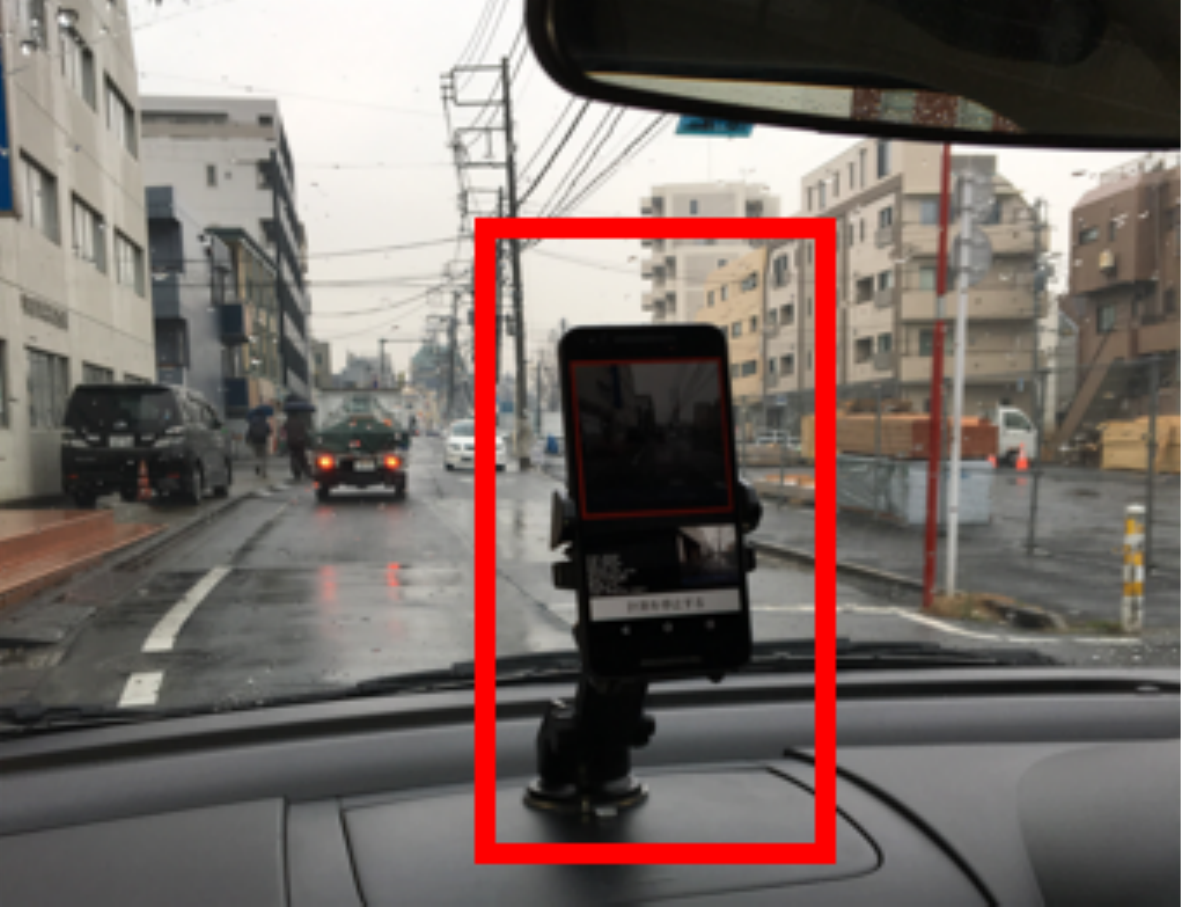}
\caption{Installation setup of the smartphone on the car. It is mounted on the dashboard of a general passenger car. Our developed application can capture a photograph of the road surface approximately 10 m ahead, which indicates that this application can photograph images while traveling on the road without leakage or duplication when the care moves at an average speed of about 40 km/h (about 10 m/s) if photographing every second. In addition, it can detect road damages in 1.5 s with high accuracy (see Section~\ref{Results}).}
\label{smartphoneInstallation}
\end{figure}

We installed a smartphone (LG Nexus 5X) on the dashboard of a car, as shown in Figure~\ref{smartphoneInstallation}, and photographed images of 600 $\times$ 600 pixels once per second. The reason we select a photographing interval of 1 s is because it is possible to photograph images while traveling on the road without leakage or duplication when the average speed of the car is approximately 40 km/h (or approximately 10 m/s). For this purpose, we created a smartphone application that can capture images of the roads and record the location information once per second; this application is also publicly available on our website.

\subsection{Data Category}
\begin{figure*}
\centering\includegraphics[width=15cm]{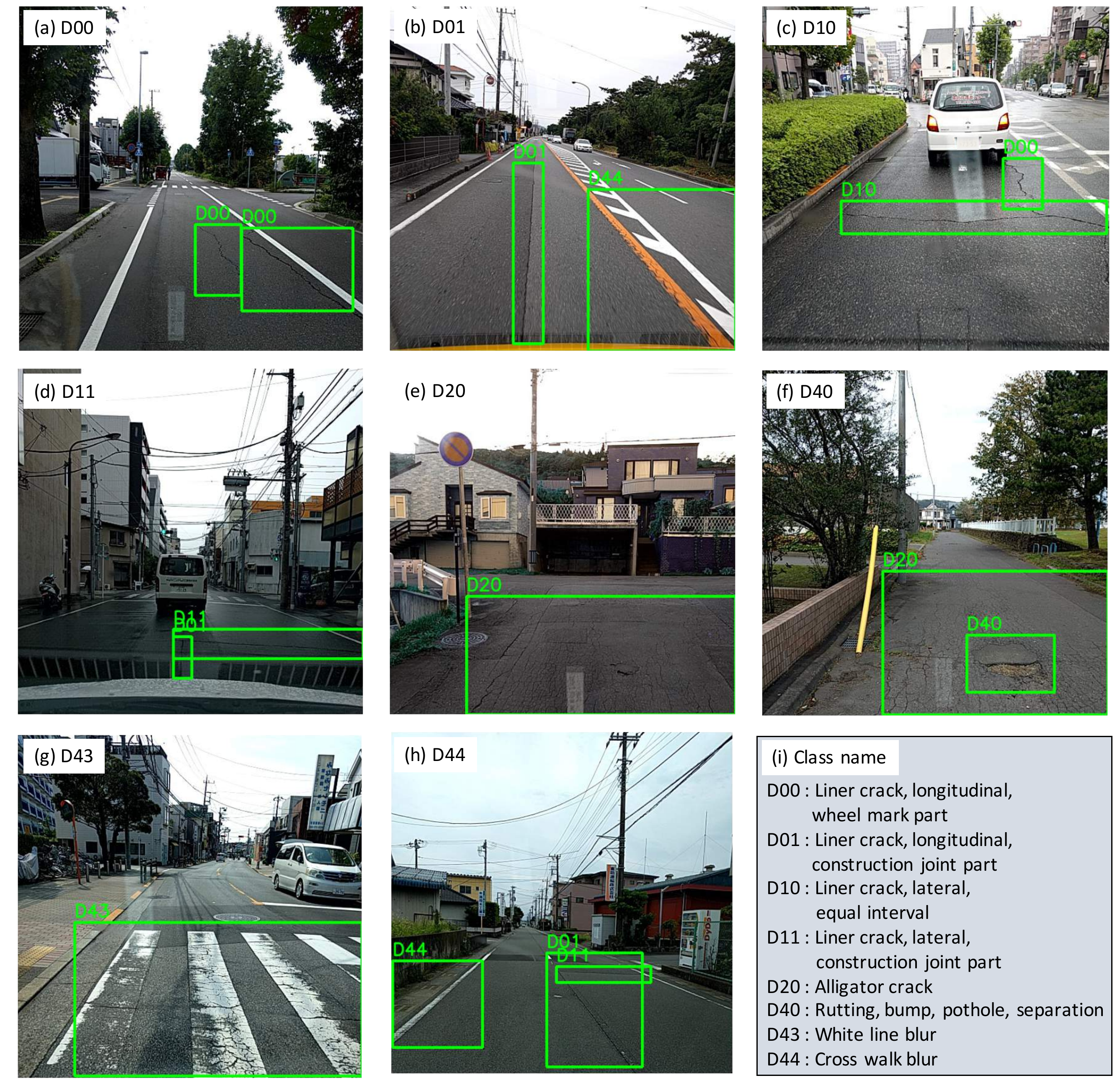}
\caption{Sample images of our dataset: (a) to (h) correspond to each one of the eight categories, and (i) shows the legend. Our benchmark contains 163,664 road images and of these, 9,053 images include cracks. A total of 9,053 images were annotated with class labels and bounding boxes. The images were captured using a smartphone camera in realistic scenarios.}
\label{sampleImages}
\end{figure*}

Table~\ref{crackTypeDef} list the different damage types and their definition. In this paper, each damage type is represented with a Class Name such as D00. Each type of damage is illustrated in the examples in Figure~\ref{sampleImages}.

As can be seen from the table, the damage types are divided into eight categories. First, the damage is classified into cracks or other corruptions. Then, the cracks are divided into linear cracks and alligator cracks (crocodile cracks). Other corruptions, include not only pot holes and rutting, but also other road damage such as blurring of white lines.

To the best of our knowledge, no previous research covers such a wide variety of road damages, especially in the case of image processing. For example, the method proposed by Jo et al.~\citep*{jo2015pothole} detects only potholes in D40, and that of Zalama et al.~\citep*{zalama2014road} classifies damage types exclusively as longitudinal and lateral, whereas the method proposed by Akarsu et al.~\citep*{akarsu2016fast} categorizes damage types into longitudinal, lateral, and alligator cracks. Further, the previous study using deep learning~\citep*{zhang2016road,maeda2016lightweight} only detects the presence or absence of damage.

\begin{table*}
\centering
\caption{Road damage types in our dataset and their definitions.}
\label{crackTypeDef}
\scalebox{0.83}{
\begin{tabular}{|ccc|c|c|}
\multicolumn{3}{c}{}                                         &\multicolumn{1}{c}{} &\multicolumn{1}{c}{} \\ \hline
\multicolumn{3}{|c|}{Damage Type}                                           &\multicolumn{1}{|c|}{Detail} &\multicolumn{1}{|c|}{Class Name} \\ \hline \hline
\multicolumn{1}{|c|}{} &  \multicolumn{1}{|c|}{}             & Longitudinal & Wheel mark part           & D00\\ \cline{4-5}
Crack                  &  \multicolumn{1}{|c|}{Linear Crack} &              & Construction joint part   & D01\\ \cline{3-5}
\multicolumn{1}{|c|}{} &  \multicolumn{1}{|c|}{}             & Lateral      & Equal interval            & D10 \\ \cline{4-5}
\multicolumn{1}{|c|}{} &  \multicolumn{1}{|c|}{}             &              & Construction joint part   & D11\\ \cline{2-5}
                       &  \multicolumn{2}{|c|}{Alligator Crack}             & Partial pavement, overall pavement & D20 \\ \hline
                       &                                     &              & Rutting, bump, pothole, separation  & D40\\ \cline{4-5}
\multicolumn{3}{|c|}{Other Corruption}                                      & White line blur                    & D43\\ \cline{4-5}
                       &                                     &              & Cross walk blur                  & D44\\ \hline
\end{tabular}
}

\footnotesize\emph{Source:} Road Maintenance and Repair Guidebook 2013~\cite{pavement01} in Japan.

\emph{Note:} In reality, rutting, bump, pothole, and separation are different types of road damage, but it was difficult to distinguish these four types using images. Therefore, they were classified as one class, viz., D40.
\end{table*}

\subsection{Data Annotation}
\begin{figure}
\centering\includegraphics[width=0.9\linewidth]{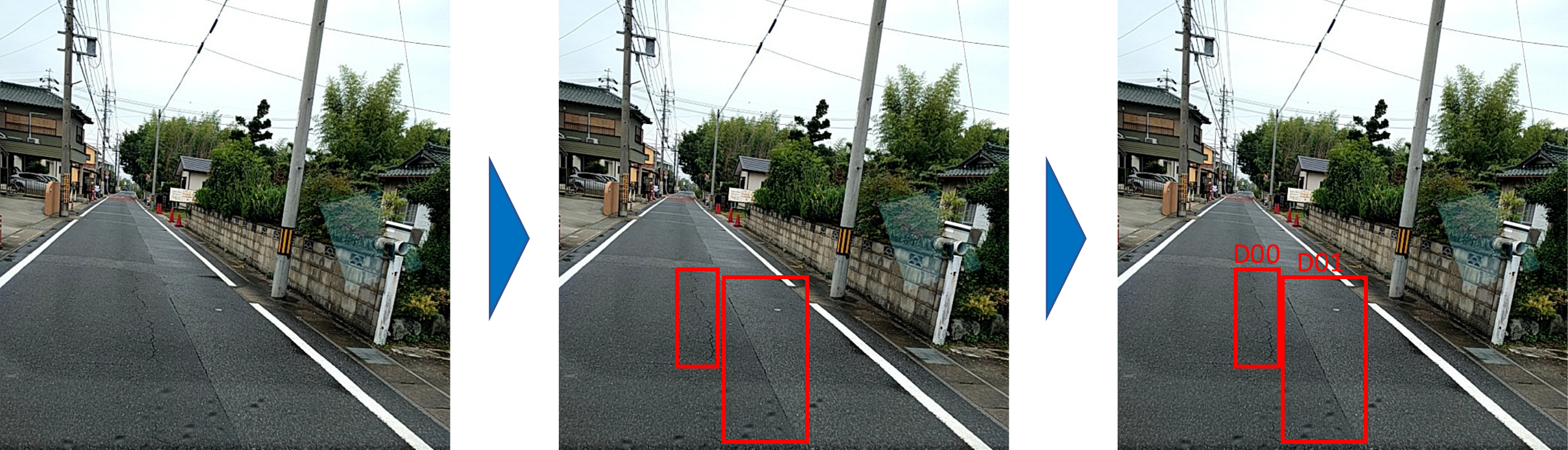}
\caption{Annotation pipeline. First, we draw the bounding box. Then, the class label is attached.}
\label{annotation_pipeline}
\end{figure}

The collected images were then annotated manually. We illustrate our annotation pipeline in Figure~\ref{annotation_pipeline}. Because our dataset format is designed in a manner similar to the PASCAL VOC~\citep*{everingham2010pascal,everingham2015pascal}, it is easy to apply it to many existing methods used in the field of image processing.

%この図の位置がtable1よりも前に来ないように

\subsection{Data Statistics}

\begin{figure*}
\centering\includegraphics[width=15cm]{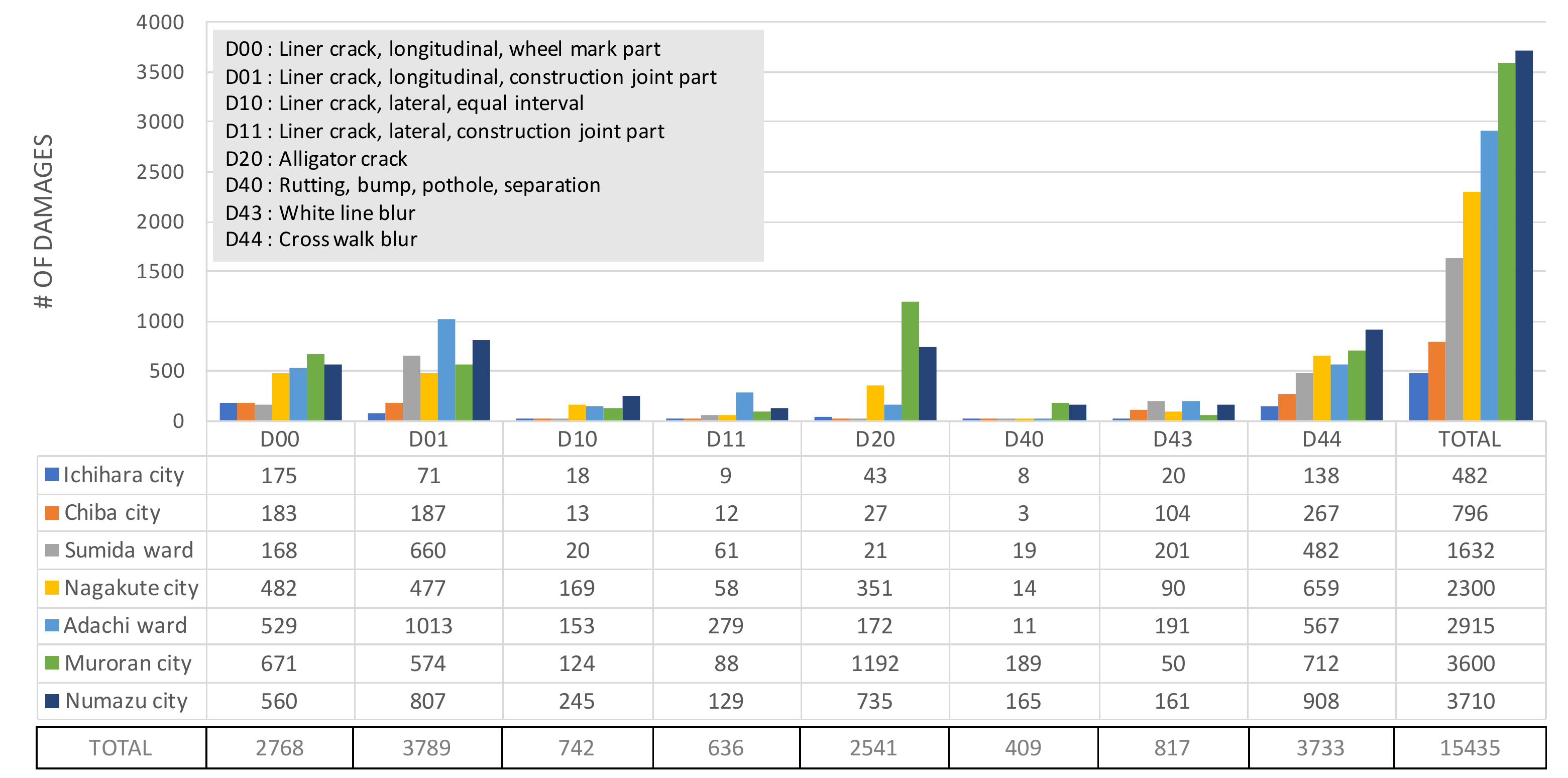}
\caption{Number of damage instances in each class in each municipality. We can see that the distribution of damage type differs for each local government. For example, in Muroran city, there are many D20 damages (1,192 damages) compared to other municipalities. This is because Muroran city is a snowy zone, therefore, alligator cracks tend to occur during the thaw of snow.}
\label{dataset_statistics}
\end{figure*}

Our dataset is composed of 9,053 labeled road damage images. Of these 9,053 images, 15,435 bounding boxes of damage are annotated. Figure~\ref{dataset_statistics} shows the number of instances per label that were collected in each municipality. We photographed a number of road images in various regions of Japan, but could not avoid biasing some of the data. For example, damages such as D40 pose a more significant danger, and therefore, road managers repair these damages as soon as they occur; thus, there are not many instances of D40 in reality. In many studies, the blurring of white lines is not considered to be damage; however, in this study, white line blur is also considered as damage.
In summary, our new dataset includes 9,053 damage images and 15,435 damage bounding boxes. The resolution of the images is 600 $\times$ 600 pixels. The area and the weather in the area are diverse, and thus, the dataset closely resembles the real world. 
We used this dataset to evaluate the damage detection model.

\subsection{Privacy Matters}
Our dataset is openly accessible by the public. Therefore, considering issues with privacy, based on visual inspection, when a person's face or a car license plate are clearly reflected in the image, they are blurred out.

\section{Experimental Setup}
\label{ExperimentalSetup}
Based on a previous study in which many neural networks and object detection methods were compared in detail~\citep*{huang2016speed}, among the state-of-the-art object detection methods, the SSD using Inception V2 and SSD using MobileNet are those with relatively small CPU loads and low memory consumption, even while maintaining high accuracy. 
However, it is important to note that the results of the abovementioned research were obtained using the COCO dataset~\citep*{lin2014microsoft}.
Because we believe that an object detection method that can be executed on a smartphone (or a small computational resource) is desirable, in this study, we trained the model using the SSD Inception V2 and SSD MobileNet frameworks.
We analyze the cases of applying the SSD using Inception and SSD using MobileNet to our dataset in detail.

\subsection{Parameter Settings}
In the object detection algorithm using deep learning, the parameters learned from the data are enormous; in addition, the number of hyper parameters set by humans is large. The parameter setting in the case of each algorithm is described below.

\subsubsection{SSD using Inception V2}
We followed the methodology mentioned in the original paper~\citep*{liu2016ssd}. The initial learning rate is 0.002, which is reduced by a learning rate decay of 0.95 per 10,000 iterations.
The input image size is 300 $\times$ 300 pixels, which indicates that the original images are resized from 600 $\times$ 600 to 300 $\times$ 300. 

\subsubsection{SSD using MobileNet}
As in the previous case, we followed the methodology mentioned in the original paper~\citep*{liu2016ssd} as well. Similar to Huang et al.~\citep*{huang2016speed}, we initialize the weights with a truncated normal distribution with a standard deviation of 0.03. The initial learning rate is 0.003 with a learning rate decay of 0.95 every 10,000 iterations. The input image size in this case is 300 $\times$ 300 pixels as well.

\subsection{Training and Evaluation}
We conducted training and evaluation using our dataset. For our experiment, the dataset was randomly divided in a ratio of 8:2; the former part was set as training data, and the latter as evaluation data. Thus, the training data included 7,240 images, and the evaluation data had 1,813 images.

\section{Results}
\label{Results}
\begin{figure*}[h]
\centering\includegraphics[width=15cm]{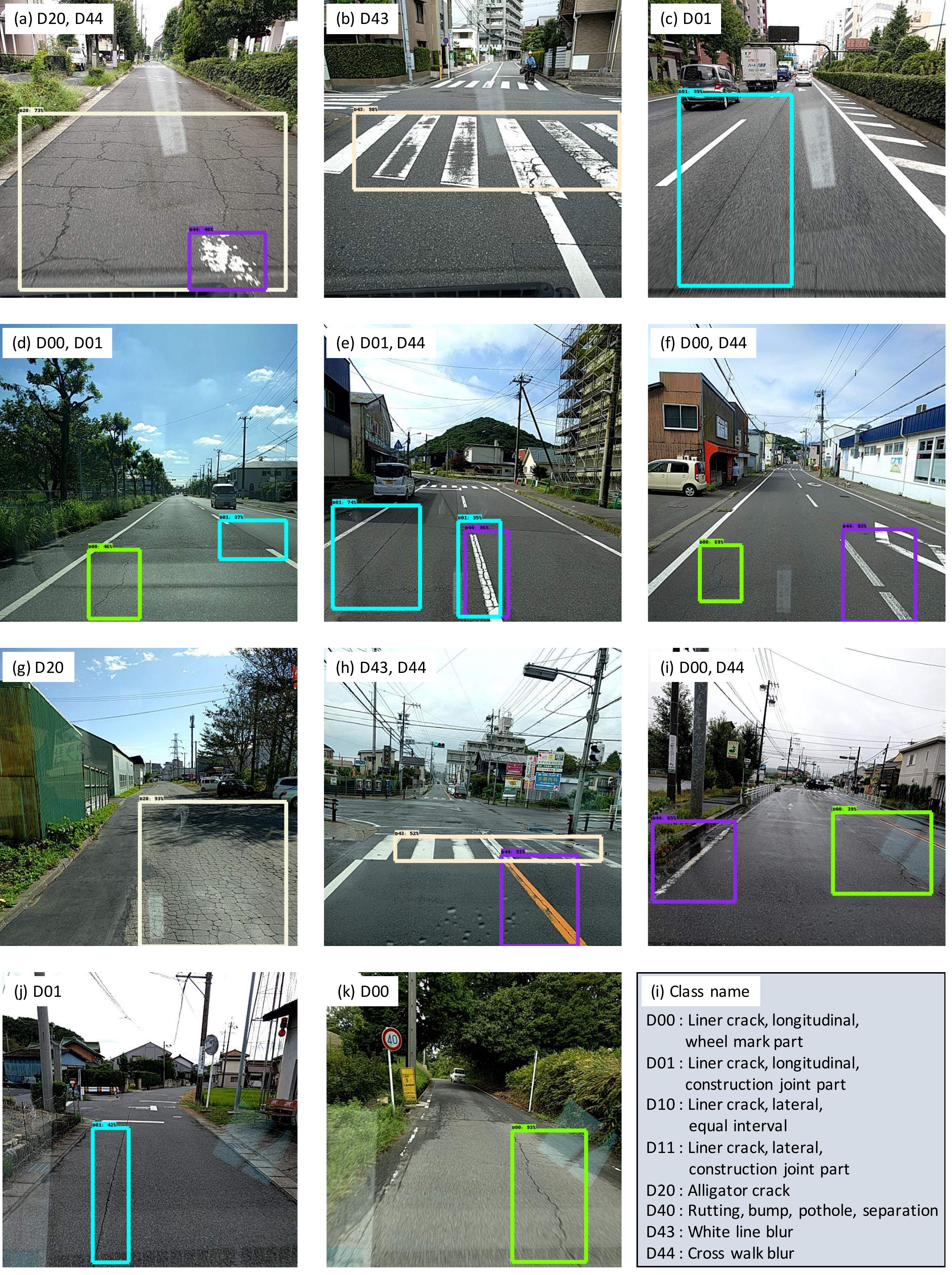}
\caption{Detected samples using the SSD MobileNet.}
\label{detectedSampleImagesMobileNet}
\end{figure*}

\begin{figure*}[h]
\centering\includegraphics[width=15cm]{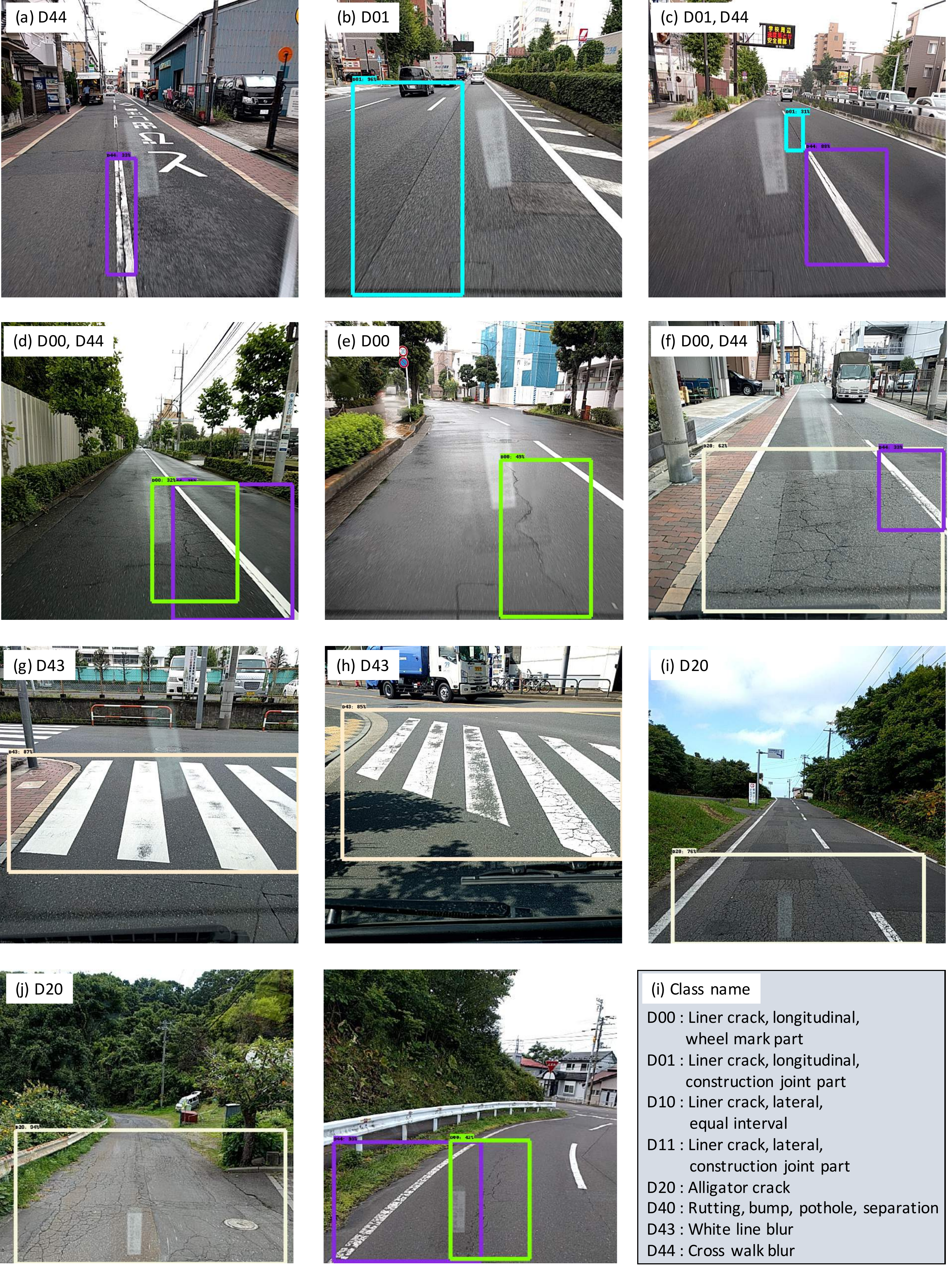}
\caption{Detected samples using the SSD Inception V2.}
\label{detectedSampleImagesInception}
\end{figure*}

In our experiment, training was performed on an PC running the Ubuntu 16.04 operating system with an NVIDIA GRID K520 GPU and 15 GB RAM memory. In the evaluation, the Intersection Over Union (IOU) threshold was set to 0.5. The detected samples are illustrated in Figures~\ref{detectedSampleImagesMobileNet} and \ref{detectedSampleImagesInception}.

We compared the results provided by the SSD Inception V2 and SSD MobileNet. These results are listed in Table~\ref{result}. Although D01 and D44 were detected with relatively high recall and precision, the value of recall is low in the case of D11 and D40; This can be attributed to the number of training data (see Figure~\ref{dataset_statistics}). 
On the contrary, D43 was detected with high recall and precision even though the number of training data is small; this is because D43 (blur of the pedestrian crossing) occupies a large proportion in the image and the feature is clear (i.e. stripped pattern). Overall, the SSD MobileNet yields better results.

\begin{figure}[h]
\centering\includegraphics[width=1\linewidth]{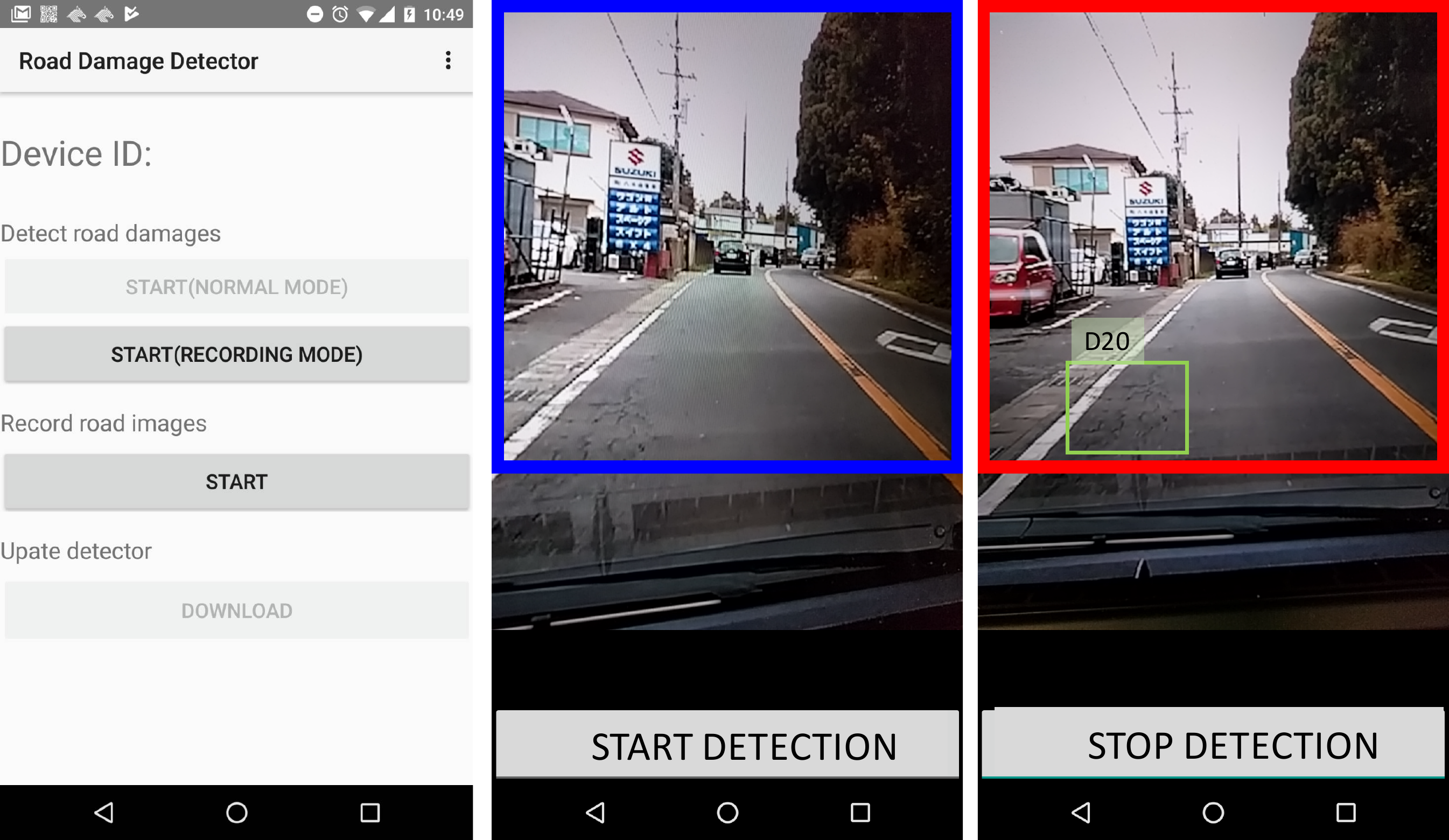}
\caption{Operating screen of our smartphone application. It is supposed to be mounted on the dashboard of a general passenger car (See Figure~\ref{smartphoneInstallation}). Detection of road surface damage is initiated by pressing the "START DETECTION" button. An image of the damaged part and the position information are transmitted to the external server only when damage is found. Using the SSD with MobileNet, this application can detect road damages (of eight types) in 1.5 s with the same accuracy as shown in Table~\ref{result}.}
\label{smartphoneApp}
\end{figure}

Next, the inference speed of each model is described in Table~\ref{runTimeSpeed}. The speed was tested on a PC with the same specifications as in the previous case and a Nexus 5X smartphone with an MSM8992 CPU and 2 RAM GB memory. In this case, the SSD Inception V2 is two times slower than the SSD MobileNet, which is consistent with the result of Huang et al.~\citep*{huang2016speed}. In addition, because the smartphone processed data in 1.5 s, when it is installed in a moving car, damage to the road surface can be detected in real time and with the same accuracy as in Table~\ref{result}. Our smartphone application, which we used to detect road damage using the trained model with our dataset (SSD with MobileNet. See Figure~\ref{smartphoneApp}) is publicly available on our website.

\begin{table*}
\caption{Detection and classification results for each class using the SSD Inception and SSD MobileNet. SIR: SSD Inception V2 Recall, SIP: SSD Inception V2 Precision, SIA: SSD Inception V2 Accuracy, SMR: SSD Recall, SMP: SSD Precision, SMA: SSD Accuracy.}
\centering
\begin{tabular}{ccccccccc}
\hline
\textbf{Class} & \textbf{D00} & \textbf{D01} & \textbf{D10} & \textbf{D11} & \textbf{D20} & \textbf{D40} & \textbf{D43} & \textbf{D44} \\
\hline
SIR & 0.22 & 0.60 & 0.10 & 0.05 & 0.68 & 0.03 & 0.81 & 0.62 \\
SIP & 0.73 & 0.84 & 0.99 & 0.95 & 0.73 & 0.67 & 0.77 & 0.81 \\
SIA & 0.78 & 0.80 & 0.94 & 0.92 & 0.85 & 0.95 & 0.95 & 0.83 \\
SMR & 0.40 & 0.89 & 0.20 & 0.05 & 0.68 & 0.02 & 0.71 & 0.85 \\
SMP & 0.73 & 0.64 & 0.99 & 0.95 & 0.68 & 0.99 & 0.85 & 0.66 \\
SMA & 0.81 & 0.77 & 0.92 & 0.94 & 0.83 & 0.95 & 0.95 & 0.81 \\
\hline
\end{tabular}{}
\label{result}
\end{table*}

\begin{table*}
\centering
\caption{Inference speed (ms) for each model for image resolution of a 300 $\times$ 300-pixel image}
\begin{tabular}{cc}
\hline
\textbf{Model Details} & Inference speed (ms)\\
\hline
SSD using Inception V2 (GPU) & 63.1 \\
SSD using MobileNet (GPU) & 30.6 \\
SSD using MobileNet (smartphone) & 1500 \\
\hline
\end{tabular}
\label{runTimeSpeed}
\end{table*}

\section{Conclusions}
\label{Conclusions}
In this study, we developed a new large-scale dataset for road damage detection and classification. In collaboration with seven local governments in Japan, we collected 163,664 road images. Then, these images with road damage were visually confirmed and classified into eight classes; out of these, 9,053 images were annotated and released as a training dataset. To the best of our knowledge, this dataset is the first one for road damage detection. We strongly believe this dataset provides a new avenue for road damage detection. In addition, we trained and evaluated the damage detection model using our dataset.  Based on the results, in the best-detectable category, we achieved recalls and precisions greater than 75\% with an inference time of 1.5 s on a smartphone. We believe that a simple road inspection method using only a smartphone will be useful in regions where experts and financial resources are lacking. To support research in this field, we have made the dataset, trained models, source code, and smartphone application publicly available. In the future, we plan to develop methods that can detect rare types of damage that are uncommon in our dataset.

\section*{Acknowledgement}
This research was supported by the National Institute of Information and Communication Technology (NICT) under contract research: ``Social Big Data Utilization R and D of Basic Technology'' (Issue D: Knowledge of the Site, Citizen\rq\ s Knowledge Organically, Development of Next-generation Citizen Collaborative Platform). 
 Additionally, we would like to express our gratitude to Ichihara city, Chiba city, Sumida ward, Nagakute city, Adachi city, Muroran city, and Numazu city for their cooperation with the experiment.



%% References


%% \bibliographystyle{model1-num-names}
\bibliographystyle{apalike}
\bibliography{sample}






% \end{thebibliography}

\end{document}